\documentclass[runningheads]{llncs}
\usepackage[T1]{fontenc}
\usepackage{graphicx}
\usepackage{subfigure}
\usepackage{subcaption} 
\usepackage{caption} 
\usepackage{hyperref} 
\usepackage{booktabs} 
\usepackage{multirow}
\usepackage{amsmath}
\usepackage{bbding}
\begin{document}
\title{IFSENet : Harnessing Sparse Iterations for Interactive Few-shot Segmentation Excellence}
\titlerunning{IFSENet}
%
\author{Shreyas Chandgothia \and Ardhendu Sekhar \and Amit Sethi}
\authorrunning{Shreyas et al.}
%
\institute{Department of Electrical Engineering, Indian Institute of Technology Bombay, Mumbai, India}
\maketitle              
\begin{abstract}
Training a computer vision system to segment a novel class typically requires collecting and painstakingly annotating lots of images with objects from that class. Few-shot segmentation techniques reduce the required number of images to learn to segment a new class, but careful annotations of object boundaries are still required. On the other hand, interactive segmentation techniques only focus on incrementally improving the segmentation of one object at a time (typically, using clicks given by an expert) in a class-agnostic manner. We combine the two concepts to drastically reduce the effort required to train segmentation models for novel classes. Instead of trivially feeding interactive segmentation masks as ground truth to a few-shot segmentation model, we propose IFSENet, which can accept sparse supervision on a single or few support images in the form of clicks to generate masks on support (training, at least clicked upon once) as well as query (test, never clicked upon) images. To trade-off effort for accuracy flexibly, the number of images and clicks can be incrementally added to the support set to further improve the segmentation of support as well as query images. The proposed model approaches the accuracy of previous state-of-the-art few-shot segmentation models with considerably lower annotation effort (clicks instead of maps), when tested on Pascal and SBD datasets on query images. It also works well as an interactive segmentation method on support images.

\end{abstract}

\section{Introduction}
Image segmentation tasks, which aim to label each pixel in an image, can be categorized into three main types: (i) Semantic segmentation assigns unique labels to each class, (ii) Instance segmentation assigns unique labels to separate objects, and (iii) Panoptic segmentation assigns both semantic class labels and instance IDs to each pixel. Unlike classification and detection, segmentation tasks require extensive manual effort to collect pixel-level annotated training data. However, even with ample high-quality data, segmentation models struggle with predicting novel classes with fewer samples effectively.

Two frameworks, few-shot segmentation and interactive segmentation, address these issues. Few-shot segmentation uses a small support set with annotated masks for an unseen class. Typically, the aim is to segment pixels of the same class in a separate query image by learning semantic feature similarity, disregarding class-specific information. Interactive segmentation involves human-in-the-loop, allowing the model to refine segmentation masks through manual clicks, scribbles, and bounding boxes.

Both Few-Shot and Interactive approaches have drawbacks. The former focuses on predicting novel target classes but relies heavily on high-quality support masks. The latter excels at accurate masks for seen classes but is limited to segmenting one image at a time, requiring a significant number of clicks for satisfactory results on \emph{every} test image.

We propose a semantic segmentation model that overcomes the limitations of few-shot and interactive learning methods while leveraging their strengths. Our model achieves simultaneous segmentation of unannotated (query set) images without pre-existing masks by accepting sparse user annotations in the form of clicks on a subset of images (support set). It generates dense segmentation masks for both the sparsely annotated images and the remaining unannotated images. Furthermore, the model can continuously improve results by iteratively incorporating more images or corrective clicks from the user on the support set.

Through iterative refinement, our model achieves query prediction results comparable to few-shot segmentation techniques without requiring densely annotated masks as support supervision. Additionally, our model achieves support prediction results on par with state-of-the-art interactive segmentation models.

\section{Related Work}
\subsection{Semantic Segmentation}
Semantic segmentation involves assigning a category to each pixel in an image. The Fully Convolutional Network (FCN) \cite{b9} introduced a paradigm shift by replacing fully-connected dense layers with convolution layers, enabling end-to-end segmentation. FCN utilizes an encoder-decoder architecture with downsampling and upsampling operations. Subsequently, improved models such as U-Net \cite{b10}, DeepLab \cite{b1}, PSPNet \cite{b21}, and HRNet \cite{b15} emerged to enhance the performance of semantic segmentation.

U-Net\cite{b10} utilizes a 'U'-shaped network architecture with systematic skip-connections. It consists of two major parts: a contracting path with convolution and pooling, and an expansive path with convolution and upsampling layers.

DeepLab \cite{b1} is a semantic segmentation model developed by Google, with progressive advancements across generations. The latest version, DeepLabv3+, introduces an enhanced ASPP module that employs parallel atrous convolutions to capture multi-scale context using multiple atrous rates. The DenseCRF post-processing module in earlier models is replaced by a more streamlined decoder module, resulting in improved segmentation refinement.

HRNet\cite{b15} introduces the concept of high-resolution representations for dense prediction tasks, such as detection and segmentation. The network incorporates multiple high-to-low-resolution convolution streams and connects them in parallel. This approach ensures the utilization of high-resolution information throughout the network.

\subsection{Few-Shot Segmentation}

Few-shot segmentation involves predicting pixel-level labels on an image, where the target classes differ from those seen during training. SG-One \cite{b20} addresses one-shot segmentation using masked average pooling on the support image and cosine similarity for query image feature relationships. PL \cite{b3} generates prototypes for each support class and employs cosine similarity to segment query images. PANet \cite{b16} introduces prototype alignment regularization for compatible embedding prototypes. CANet \cite{b19} employs a two-branch dense comparison module to compare multi-level features between support and query images.

PFENet \cite{b14} is a few-shot semantic segmentation model that enhances generalization on unseen classes. It achieves spatial consistency between query and support targets and effectively utilizes mid-level semantic information of training classes. The authors propose a training-free prior mask generation method, improving generalization. They also introduce a multi-scale architecture that addresses spatial inconsistency by enriching query features with support features and prior masks in an adaptive manner.

\subsection{Interactive Segmentation}

Interactive segmentation allows users to segment images by providing object cues, facilitating human-computer interaction. It offers a faster and more convenient alternative to manual pixel labeling for generating ground-truth masks. Grabcut \cite{b11} introduced an iterative energy minimization approach using a Gaussian mixture cost function. The first CNN-based interactive segmentation model, DIOS \cite{b18}, proposed a click simulation strategy later utilized by other authors. CNN-based architectures such as \cite{b5}, \cite{b6}, \cite{b7}, \cite{b8}, emphasized attention mechanisms. Other interactive feedback forms, like scribbles and bounding boxes, were also explored in works like \cite{b2}, \cite{b11}, and \cite{b17}.

RITM \cite{b13} is a click-based interactive segmentation model that incorporates segmentation masks from previous steps. Its architecture is similar to semantic segmentation models, using DeepLabv3+ with Resnet backbone and HRNet+OCR semantic segmentation architectures. Additional input channels encode spatial information of positive and negative clicks, which are combined with the backbone's output using element-wise addition. RITM samples points from the largest mislabelled region between ground truth and prediction, supporting iterative training. It also allows the optional inclusion of the output mask from a previous iteration, enhancing the power of iterative learning.

\section{Method, Training, and Validation}

Given a dataset of images, IFSENet generates binary segmentation masks, separating pixels belonging to a specific semantic class (e.g., person, car, horse) as foreground and the remaining image as background. It can handle novel classes unseen during training. The user provides information about the class to segment by giving positive and negative clicks on a subset of the images.

The images on which the user provides clicks become part of the support set $S$ and the rest of the images constitute the query set $Q$. The model generates segmentation masks for both sets. Through an iterative process, the user can refine the masks by providing clicks and incorporating the predicted masks from previous iterations. This loop allows for continuous improvement of the segmentation results.

\begin{figure*}
    \centering
    \includegraphics[width=0.95\textwidth]{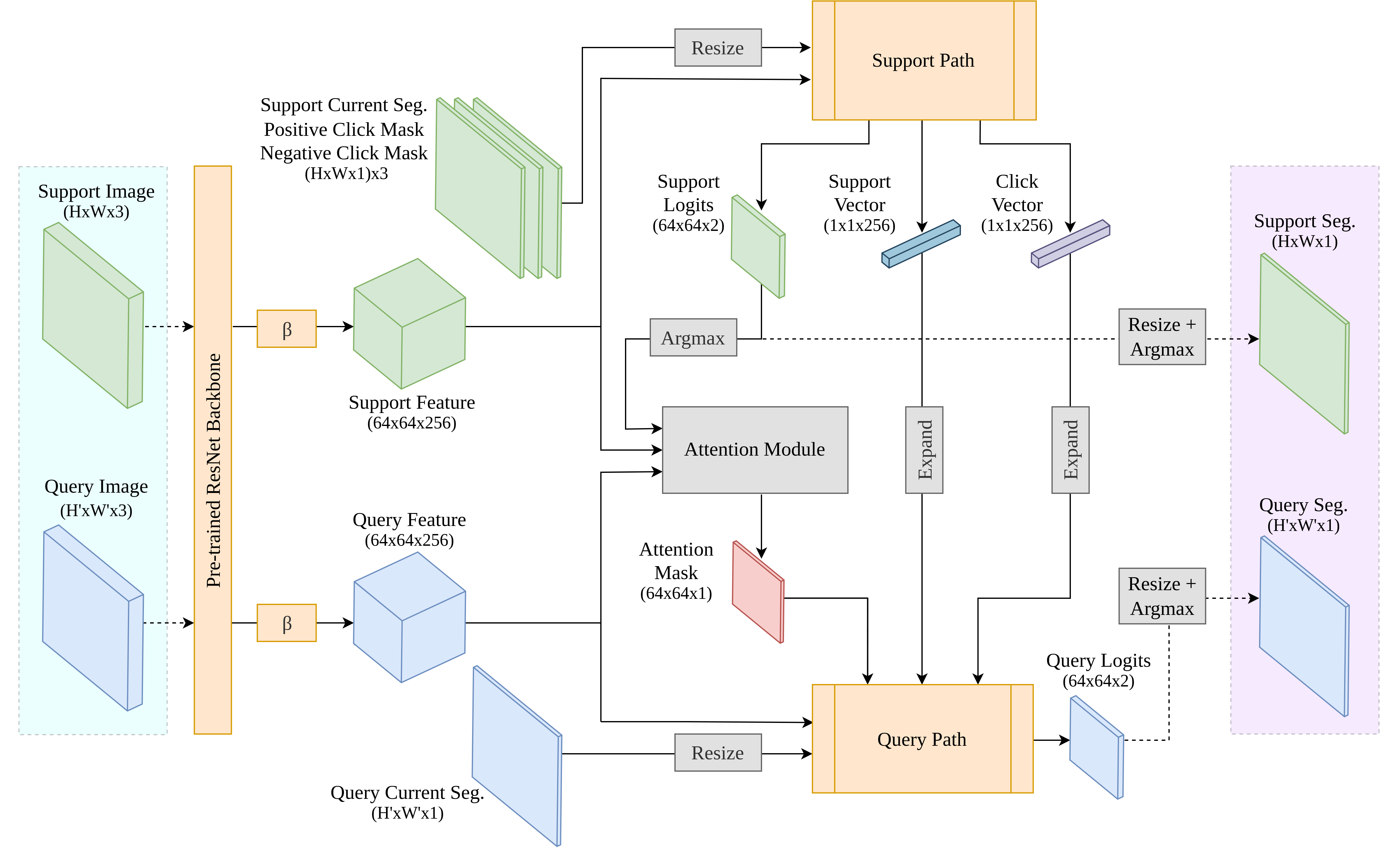}
\caption{Architecture of IFSENet: Notion-wise yellow blocks are operations with learnable parameters, grey blocks are training-free operations, $\beta$ block is 1x1conv+RELU, resize block is spatial bilinear interpolation, argmax block operates along the channel dimension, expand block makes multiple copies of a 1x1xC vector and stacks them to make the desired spatial dimension.}
    \label{fig:IFS_architecture}
\end{figure*}
\begin{table*}
    \centering
    \begin{tabular}{c|c|c|c|c}
    \toprule[1.0pt]
    \multirow{2}{*}{Method} & \multicolumn{3}{c}{ Input } & \multicolumn{1}{c}{ Output }\\
    & Support Image & Support Mask & Support Clicks & Query Mask Prediction \\
    \toprule[1.0pt]
    GFSSM & \Checkmark & \Checkmark & \text{\sffamily X} & \Checkmark  \\
    IFSENet(Ours) & \Checkmark & \text{\sffamily X} & \Checkmark & \Checkmark  \\
    \toprule[1.0pt]
    \end{tabular}
    \caption{Comparison of methods of General Few Shot Segmentation Models(GFSSM) with our proposed IFSENet. Genreally, few shot segmentation models use support images and their dense support masks to predict the query masks for query images. But IFSENet needs support clicks instead of support masks.}
    \label{table:method_comparison}    
\end{table*}
The model's architecture, depicted in Figure \ref{fig:IFS_architecture}, focuses on learning the similarities between support and query images and leveraging the importance of clicks on the support image for supervision. During training, the model avoids learning class-specific information as it aims to predict novel classes during validation, relying on user clicks.

Both the support and query images undergo feature extraction using a pre-trained ResNet backbone, originally trained on ImageNet classes. The support feature is then processed through the support path, incorporating click masks and an optional previous prediction mask, to produce a new segmentation mask for the support image. This part of the network resembles an interactive segmentation architecture. Similarly, the query feature goes through the query path, accompanied by the optional previous prediction mask, to generate an updated mask. Since direct supervision is absent for the query image, information from the support path needs to be propagated. Thus, the query path receives three additional inputs: a support vector and a click vector from the support path, as well as an attention mask from the attention module.

It should be noted that the network is capable of handling multiple support and query images. Handling multiple query images is straightforward as each prediction is independent; so the model processes them one by one to generate their respective outputs. Similarly, for multiple support images, predicting their outputs from the support path is trivial. However, it is necessary to propagate the cumulative information from the multiple support images to the query path. If there are $k$ support images, the support path produces a total of $k$ support vectors and $k$ click vectors, while the attention module generates $k$ attention masks. To combine these outputs, we simply average across the $k$ of them and pass them to the query path.

\begin{figure*}
    \centering
    \includegraphics[width=0.95\textwidth]{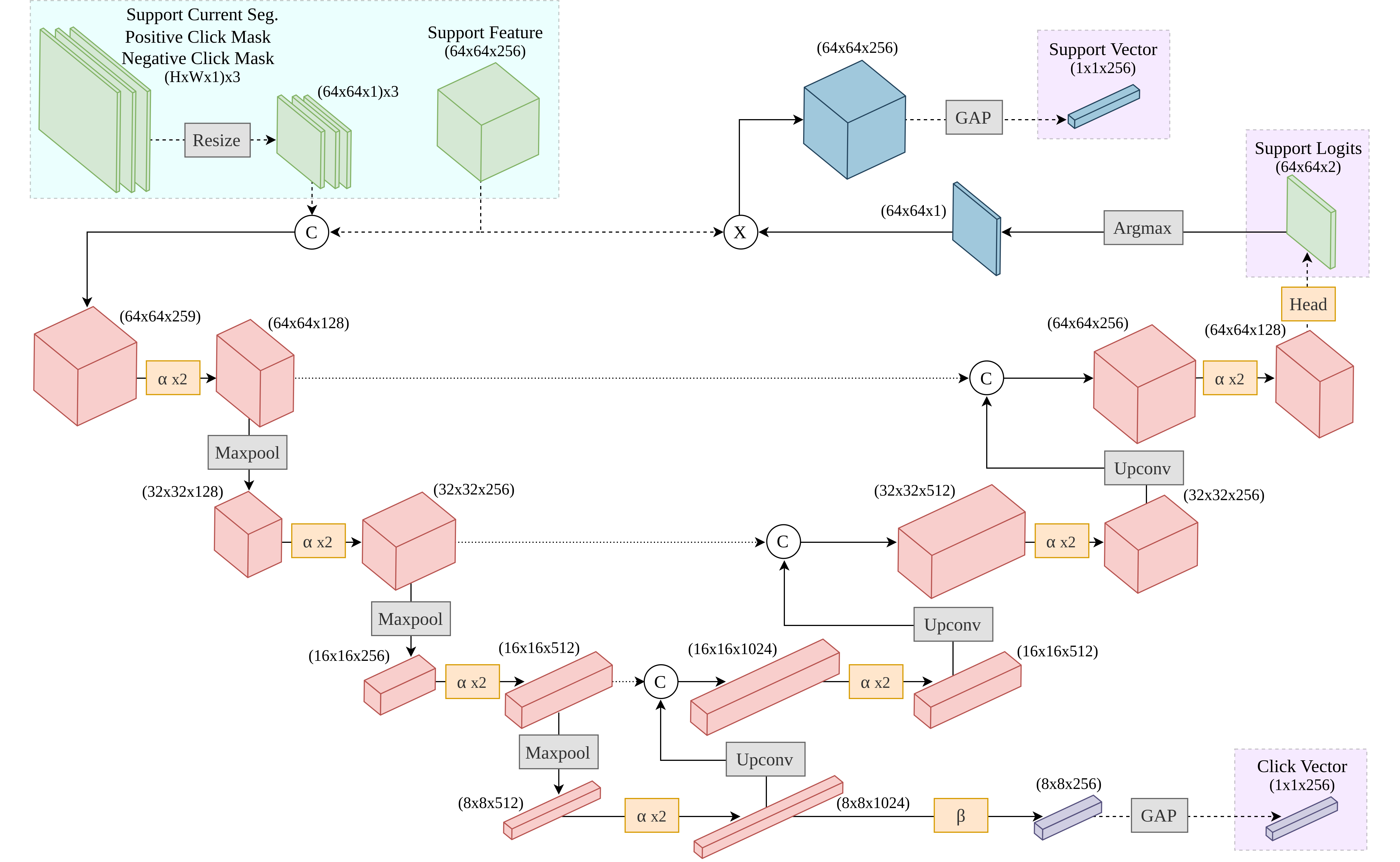}
    \caption{Architecture of the support (see Figure~\ref{fig:IFS_architecture}): Notation-wise `C' is channel concatenation, $\alpha$ block is 3x3conv+RELU, $\beta$ block is 1x1conv+RELU, max pool halves the spatial dimensions, upconv block doubles the spatial dimensions but also halves the channel dimension, GAP block is spatial global average pooling operation, head block is 1x1conv with 2-channel output.}
    \label{fig:IFS_support_path}
\end{figure*}

\subsection{Support path}

The support path architecture of our network is shown in Figure \ref{fig:IFS_support_path}. It follows a U-Net styled structure, having a contracting half that shrinks the feature spatial dimension 8x times, and an expanding half that restores the original spatial dimensions. We augment the feature map with three additional channels, i.e. two for click masks and one for the previously available segmentation mask, if available. Apart from the support image prediction in the form of logits, this part of the network produces two additional outputs:

\noindent\textbf{Support vector}: Contains the information of the foreground pixels in the support image, which later serves as supervision for the query. To obtain this vector, the binary support prediction, calculated from the support logits, is broadcasted and multiplied channel-wise with the support feature, followed by a GAP operation. 

\noindent\textbf{Click Vector:} To propagate click information from the support to the query path, we cannot pass the click masks alone as they lack the context of the support image features. Instead, we utilize the output of the last layer of the encoder, as it contains both the support image and click information. To align the dimensions with the support vector, we reduce the channel dimensions by applying a 1x1 convolution followed by a GAP operation.

\subsection{Attention Module}

The attention module takes the support features, query features, and the output logits from the support path as inputs. Its primary function is to generate an attention mask for the query image, which signifies the pixel-wise correspondence between the support and query images. In other words, it determines which pixels in the query image are most similar to the foreground pixels in the support image. This attention mask is obtained by calculating the cosine similarity between the support and query features on a pixel-wise basis. A higher value in the attention mask indicates a stronger similarity between the query pixel feature and at least one foreground support pixel feature. By leveraging this attention mechanism, the model can effectively align and relate the relevant information between the support and query images, facilitating accurate segmentation.

\begin{figure*}
    \centering
    \includegraphics[width=1.0\textwidth]{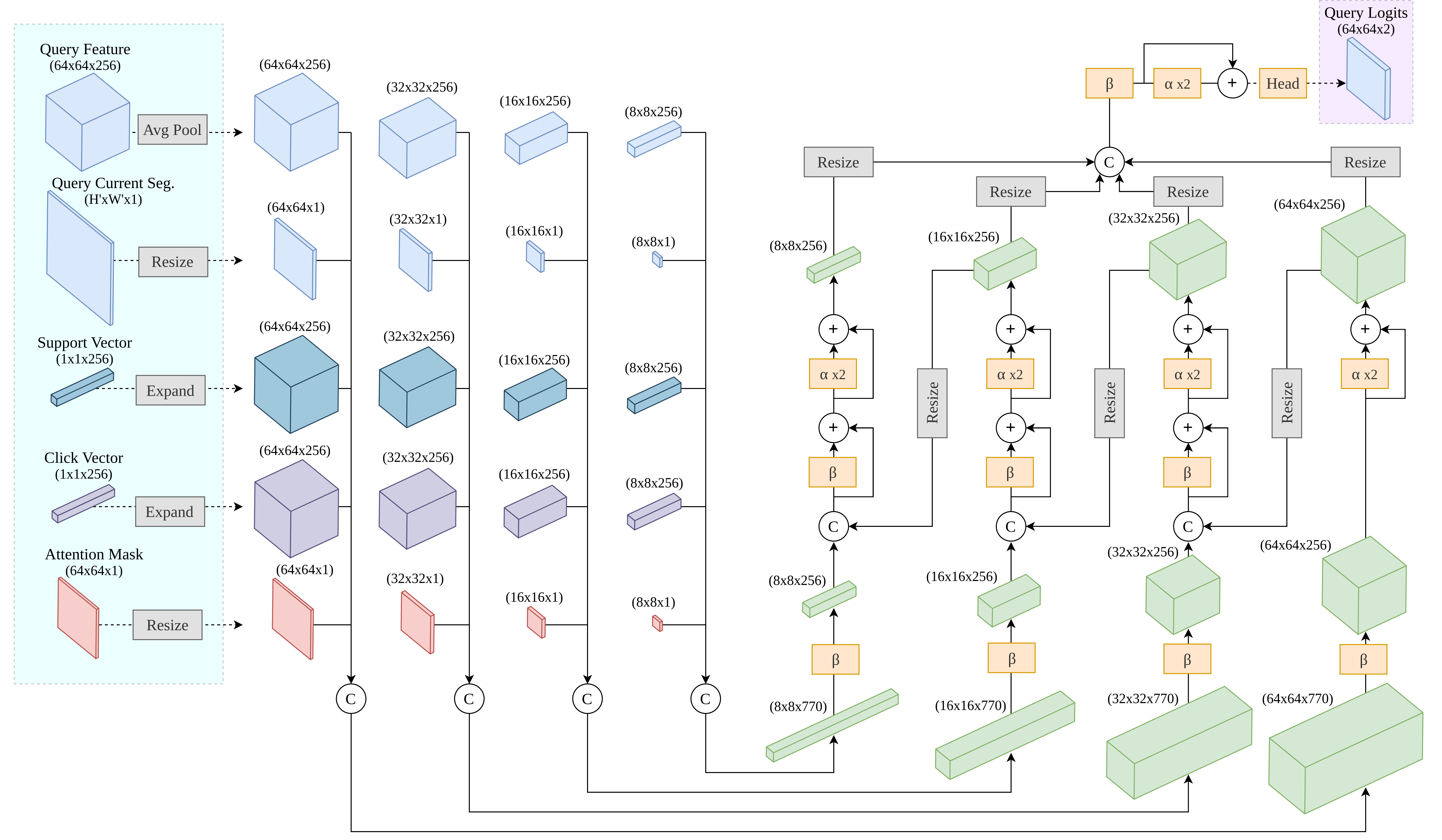}
    \caption{Architecture of the query path (see Figure~\ref{fig:IFS_architecture}): Notation-wise `C' is channel concatenation, $\alpha$ block is 3x3conv+RELU operations, $\beta$ block is 1x1conv+RELU operations, head block is 1x1conv operation with 2-channel output.}
    \label{fig:IFS_query_path}
\end{figure*}

\subsection{Query path}

The query path architecture of our network (Figure \ref{fig:IFS_query_path}) utilizes a multi-scale parallel processing approach inspired by PFENet's Feature Enrichment Module \cite{b14}. This allows for adaptive enrichment of query features with support vectors, click vectors, and attention masks. Inputs are resized, expanded, and concatenated before being processed by convolutional layers. Parallel processing layers at different scales produce multiple outputs, which are then fused and post-processed to obtain the final query prediction.

\subsection{Dataset: Pascal + SBD}

The Pascal Visual Object Classes (VOC) dataset is widely used in object detection, semantic segmentation, and classification tasks. It comprises 17,125 images, with 2,913 having segmentation masks for 20 object classes. The Semantic Boundaries Dataset (SBD) is a complementary dataset with 11,355 images that share the same 20 class labels as Pascal VOC. We merge the images and annotations from both datasets, prioritizing the higher-quality segmentation masks from SBD for common images. The merged dataset consists of 12,031 image-mask pairs. For few-shot segmentation, we split the dataset into training and validation classes, with 5 classes reserved for validation and 15 classes used for training. The dataset is denoted as Pascal-$5^i$ having four folds in total.

\subsection{Training Regime}

To begin with, we select the subset of images from our dataset, which contain at least 1 training class object within them. We also select the number of shots \textit{\textbf{k}}. Then we run a training loop for a chosen number of epochs, where each epoch involves iterating over all the images once. Below are the steps followed for each image in each epoch:

\noindent 1. Read the current image as query: (\textit{\textbf{q\_image}})
    
\noindent 2. Randomly select a target class from the query image, in case multiple classes are present. (\textit{\textbf{class\_chosen}})
    
\noindent 3. Obtain the binary ground truth mask corresponding to the \textit{\textbf{class\_chosen}} for the \textit{\textbf{q\_image}}. (\textit{\textbf{q\_mask}})
    
\noindent 4. Obtain the stored predicted query mask on this \textit{\textbf{q\_image}} and this \textit{\textbf{class\_chosen}} from a previous iteration, if it exists, else initialize as a blank mask. (\textit{\textbf{q\_currseg}})
    
\noindent 5. Randomly select \textit{\textbf{k}} images having an object from the \textit{\textbf{class\_chosen}} present within them and make them the support. (\textit{\textbf{s\_images}})
    
\noindent 6. Obtain their binary ground truth masks corresponding to the \textit{\textbf{class\_chosen}}. (\textit{\textbf{s\_masks}})
    
\noindent 7. Obtain the stored predicted support masks on these \textit{\textbf{s\_images}} and this \textit{\textbf{class\_chosen}} from previous iterations, if they exist, else initialize with blank masks. (\textit{\textbf{s\_currsegs}})
    
\noindent 8. Obtain the stored positive and negative click masks for these \textit{\textbf{s\_images}} and this \textit{\textbf{class\_chosen}} from previous iterations, if they exist, else initialize as blank masks. (\textit{\textbf{s\_posclicks}} and \textit{\textbf{s\_negclicks}})

\noindent 9. Provide \textit{\textbf{s\_images}}, \textit{\textbf{s\_currsegs}}, \textit{\textbf{s\_posclicks}}, \textit{\textbf{s\_negclicks}} \textit{\textbf{q\_image}} and \textit{\textbf{q\_currseg}} as input to the model.
    
\noindent 10. Obtain the segmentation outputs and update the stored versions of \textit{\textbf{q\_currseg}} and \textit{\textbf{s\_currsegs}} with the new predictions for future iterations.

\textbf{Implementation Details:} Our models are trained on Pascal-$5^i$ for 100 epochs with learning rate 0.0025 and batch size 4. We use SGD as our optimizer. Momentum and weight decay are set to 0.9 and 0.0001 respectively. We adopt the `poly’ policy to decay the learning rate. During training, samples are processed with mirror operation, random rotation from -10 to 10 degrees and random crop of size 512x512. During evaluation, each input sample is resized to the training patch size but with respect to its original aspect ratio by padding zeros. We continuously save the query and support predictions as well as click masks. During the next iteration of the same training image, we use the previously saved masks with probability 0.9 or reset those to blank masks with probability 0.1. Note that the ground truth support and query masks are only required for loss calculation purposes and are not given as inputs to the model.

\subsection{Validation Regime}

For this part, we work with only those images, which contain at least 1 validation class object within them. The validation proceeds in an episodic fashion, where each episode begins by selecting a \textit{\textbf{class\_chosen}}, number of support images \textit{\textbf{s}}, and number of query images \textit{\textbf{q}}. The episode involves iteratively providing clicks on the support images and updating all the predicted segmentation. Below are the steps followed for each episode:

\noindent 1. Randomly select \textit{\textbf{s}} images as support, having an object from the \textit{\textbf{class\_chosen}} present within them (\textit{\textbf{s\_images}})

\noindent 2. Randomly select \textit{\textbf{q}} images as query, having an object from the \textit{\textbf{class\_chosen}} present within them (\textit{\textbf{q\_images}})

\noindent 3. Obtain their binary ground truth mask corresponding to the \textit{\textbf{class\_chosen}} (\textit{\textbf{s\_masks}} and \textit{\textbf{q\_masks}})

\noindent 4. Initialize \textit{\textbf{s\_currsegs}}, \textit{\textbf{s\_posclicks}}, \textit{\textbf{s\_negclicks}} and \textit{\textbf{q\_currsegs}} as blank masks

\noindent 5. Iterate for 20 clicks:

i. Add one positive or negative click to each support image from \textit{\textbf{s\_images}}, based on the largest error region between the respective \textit{\textbf{s\_mask}} and \textit{\textbf{s\_currseg}}.

ii. Store the new click for each support image in the respective click masks \textit{\textbf{s\_posclicks}} or \textit{\textbf{s\_negclicks}}.

iii. Provide \textit{\textbf{s\_images}}, \textit{\textbf{s\_currsegs}}, \textit{\textbf{s\_posclicks}}, \textit{\textbf{s\_negclicks}} \textit{\textbf{q\_images}} and \textit{\textbf{q\_currsegs}} as input to the model. 

iv. Obtain the segmentation outputs on all the \textit{\textbf{s+q}} images and update their stored versions \textit{\textbf{s\_currsegs}} and \textit{\textbf{q\_currsegs}} with the new predictions.
        
\textbf{Implementation Details:} For generating the validation results, we conduct 100 episodes for each of the classes available in the dataset. We fix the number of query images per episode to \textit{\textbf{q}}=5 and use two different values for the number of support images i.e. \textit{\textbf{s}}=1 and \textit{\textbf{s}}=5. Note that the ground truth support and query masks are required for metric evaluation purposes and automating the clicking process, but are not provided as inputs to the model.

\subsection{Click Sampling Strategy}

\textbf{Training:} Positive clicks are sampled randomly from the ground truth foreground (0.2) and false negative regions (0.8). Negative clicks are sampled randomly from the ground truth background (0.04), other class objects in the background (0.06), border region around the ground truth foreground (0.1), and false positive regions (0.8). If any mask is blank, probabilities are redistributed. Higher weight is given to false positive and false negative regions to simulate validation behavior.

\textbf{Validation:} We simulate user click behavior by identifying the largest mislabeled region between the ground truth and current prediction. A click is then provided near the center of this region. If the selected pixel is a false negative, a positive click is given; otherwise, a negative click is provided.

\subsection{Evaluation Metrics}

IoU is used to evaluate general semantic segmentation by measuring the overlap between masks. Values range from 0 to 1, with higher values indicating better overlap. For few-shot tasks, Class mIoU averages foreground IoU across validation classes. In interactive segmentation, mIoU is the average foreground IoU across validation images, regardless of classes.

\subsection{Loss Function}

We use simple pixel-level Binary Cross Entropy (BCE) loss. The total loss is the weighted sum of the losses over the $k$ support image predictions ($\mathcal{L}_{S}^i$), losses over the $n$ intermediate query predictions at the different scales ($\mathcal{L}_{Q,1}^i$) and the final query prediction ($\mathcal{L}_{Q,2}$).

$$\mathcal{L}= \frac{1}{k}\sum_{i=1}^k \mathcal{L}_{S}^i + \frac{1}{n}\sum_{i=1}^n \mathcal{L}_{Q,1}^i + \mathcal{L}_{Q,2}$$

\section{Experiment and Results}

\begin{figure*}
    \centering
         \includegraphics[width=0.95\textwidth]{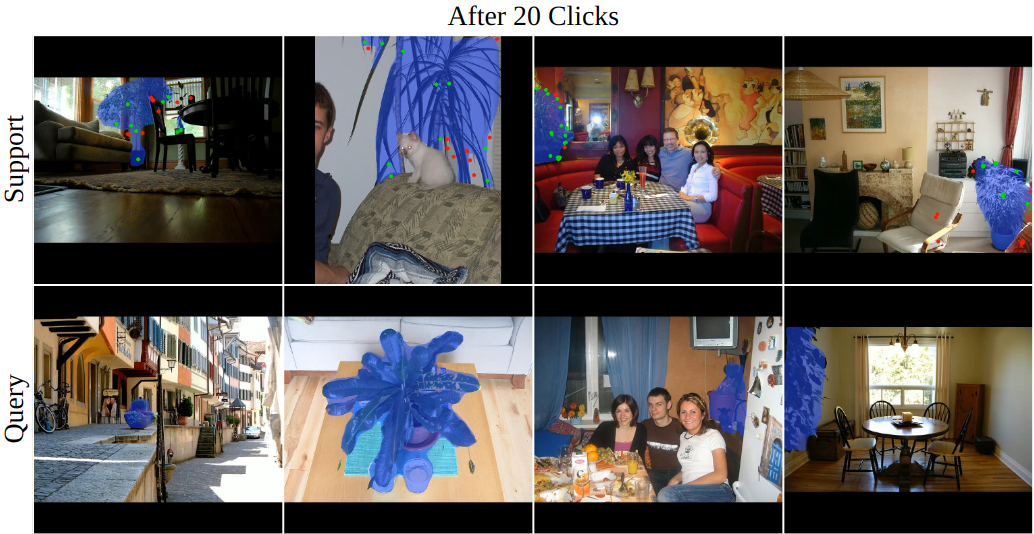}
         \caption{Visualization of segmentation episodes on potted plants class : Positive clicks are green dots, negative clicks are red dots in the support images, and the segmentation masks are overlaid blue regions on both support and query images.}
         \label{fig:visual_results}
\end{figure*}

The qualitative results for selected episodes on validation classes are shown in Figure~\ref{fig:visual_results}. In the below figures, the network initially mis-segments the `Person' class for the `Potted Plant' class but corrects it with additional support image clicks. No clicks are given on the query images in our test results, but users have the option to add clicks for poor-performing queries and transfer them to the support set. Demo videos of our results are hosted at \href{https://drive.google.com/drive/folders/18P8vXBJVGbSHKCVFNcJDwncBMDxhXLV6?usp=sharing}{this} link.
\begin{table*}

    \centering
    \begin{tabular}{ccccccccccc}
    
    \toprule[1.0pt]
    
    \multirow{2}{*}{Method} & \multicolumn{5}{c}{ 1 Shot } & \multicolumn{5}{c}{ 5 Shot }\\
    
    \cmidrule[1.25pt](l{0.5em}r{0.5em}){2-6}\cmidrule[1.25pt](l{0.5em}){7-11}
    
    & Fold-0 & Fold-1 & Fold-2 & Fold-3 & Mean & Fold-0 & Fold-1 & Fold-2 & Fold-3 & Mean \\
    
    \toprule[1.0pt]
    
    SG-One\cite{b20} & 40.2 & 58.4 & 48.4 & 38.4 & 46.3 & 41.9 & 58.6 & 48.6 & 39.4 & 47.1 \\
    
    PANet\cite{b16} & 42.3 & 58.0 & 51.1 & 41.2 & 48.1 & 51.8 & 64.6 & \textbf{59.8} & 46.5 & 55.7 \\
    
    CANet\cite{b19} & 52.5 & 65.9 & \underline{51.3} & \underline{51.9} & 55.4 & 55.5 & 67.8 & 51.9 & 53.2 & 57.1\\
    
    PFENet\cite{b14} & \textbf{61.7} & \textbf{69.5} & \textbf{55.4} & \textbf{56.3} & \textbf{60.8} & \underline{63.1} & \textbf{70.7} & 55.8 & \textbf{57.9} & \textbf{61.9} \\
    
    Ours & \underline{59.4} & \underline{66.5} & 50.3 & 51.5 & \underline{56.9} & \textbf{64.3} & \underline{70.1} & \underline{55.9} & \underline{55.7} & \underline{61.5} \\
    
    \toprule[1.0pt]
    
    \end{tabular}
    
    \caption{\textbf{Main result:} An unfair comparison of our model \emph{using only 20 clicks per support image and no query clicks} with few-shot segmentation techniques \emph{that use dense ground truth masks} on class mIoU for query images of validation classes from Pascal-$5^i$. Best results are bold, second-best results are underlined.}

    \label{table:query_results}
    
\end{table*}
\begin{figure*}
    \centering
         \includegraphics[width=0.95\textwidth]{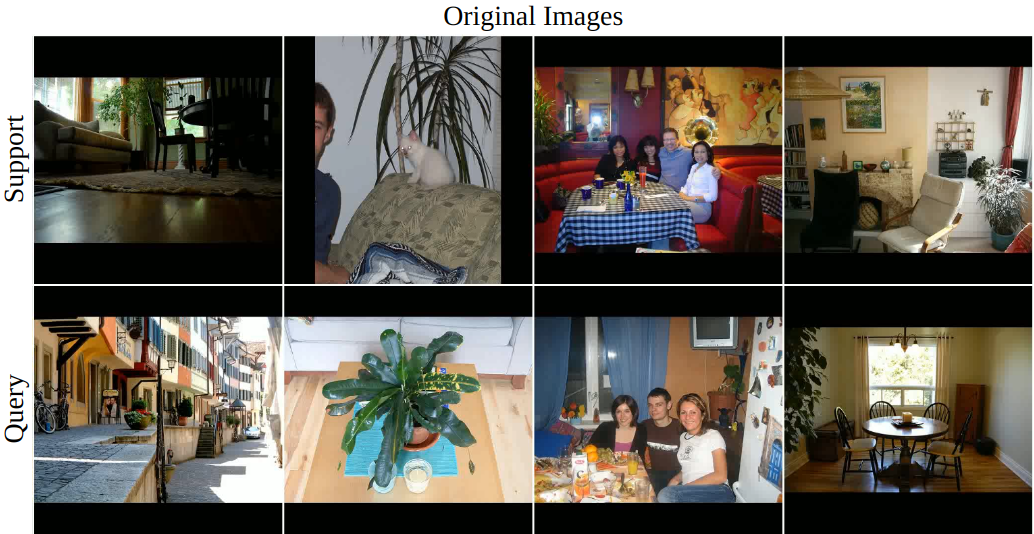}
\end{figure*}

\begin{figure*}
    \centering
         \includegraphics[width=0.95\textwidth]{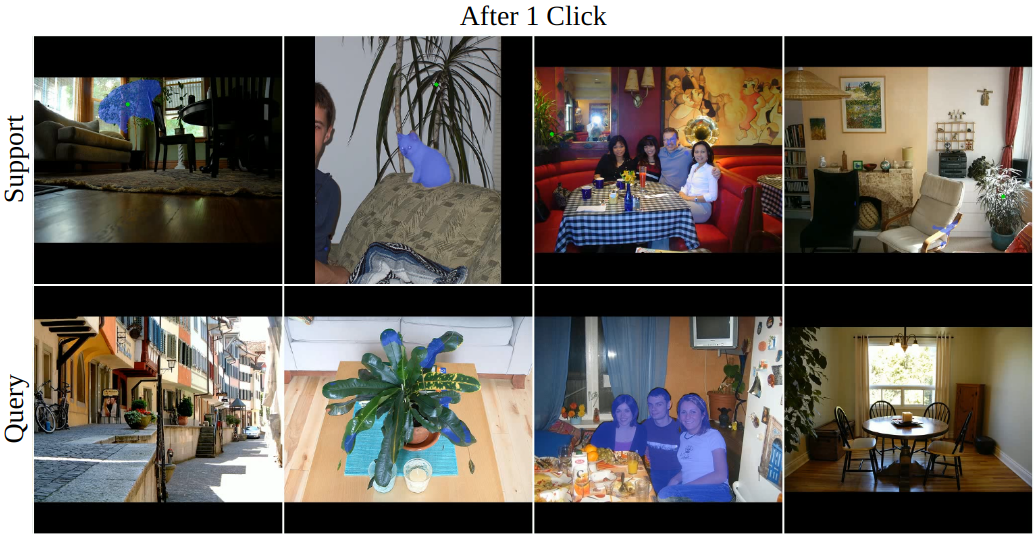}
\end{figure*}

\begin{figure*}
    \centering
         \includegraphics[width=0.95\textwidth]{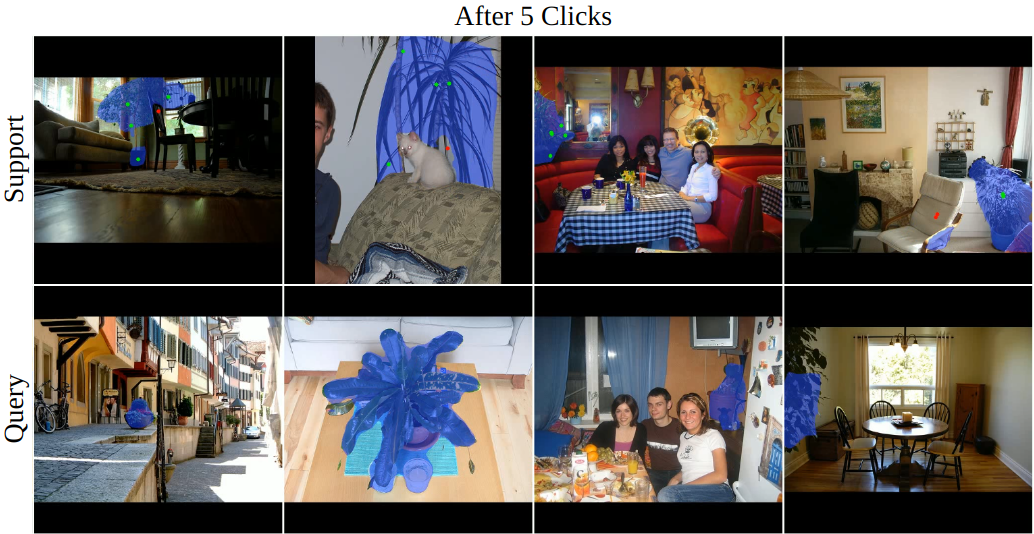}
\end{figure*}

\begin{figure}[!ht]
    \centering
    \includegraphics[width=1.0\linewidth]{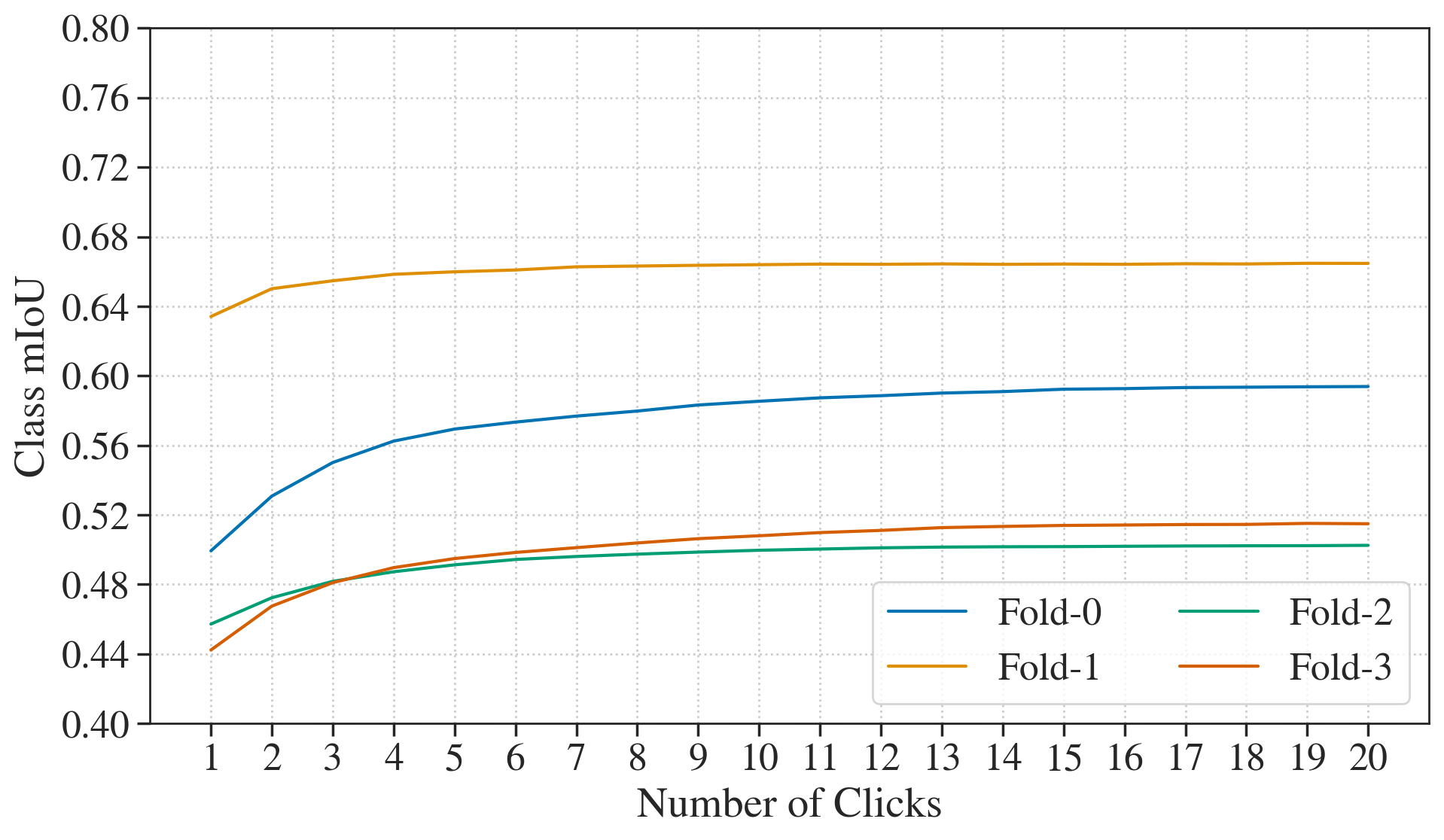}
    \includegraphics[width=1.0\linewidth]{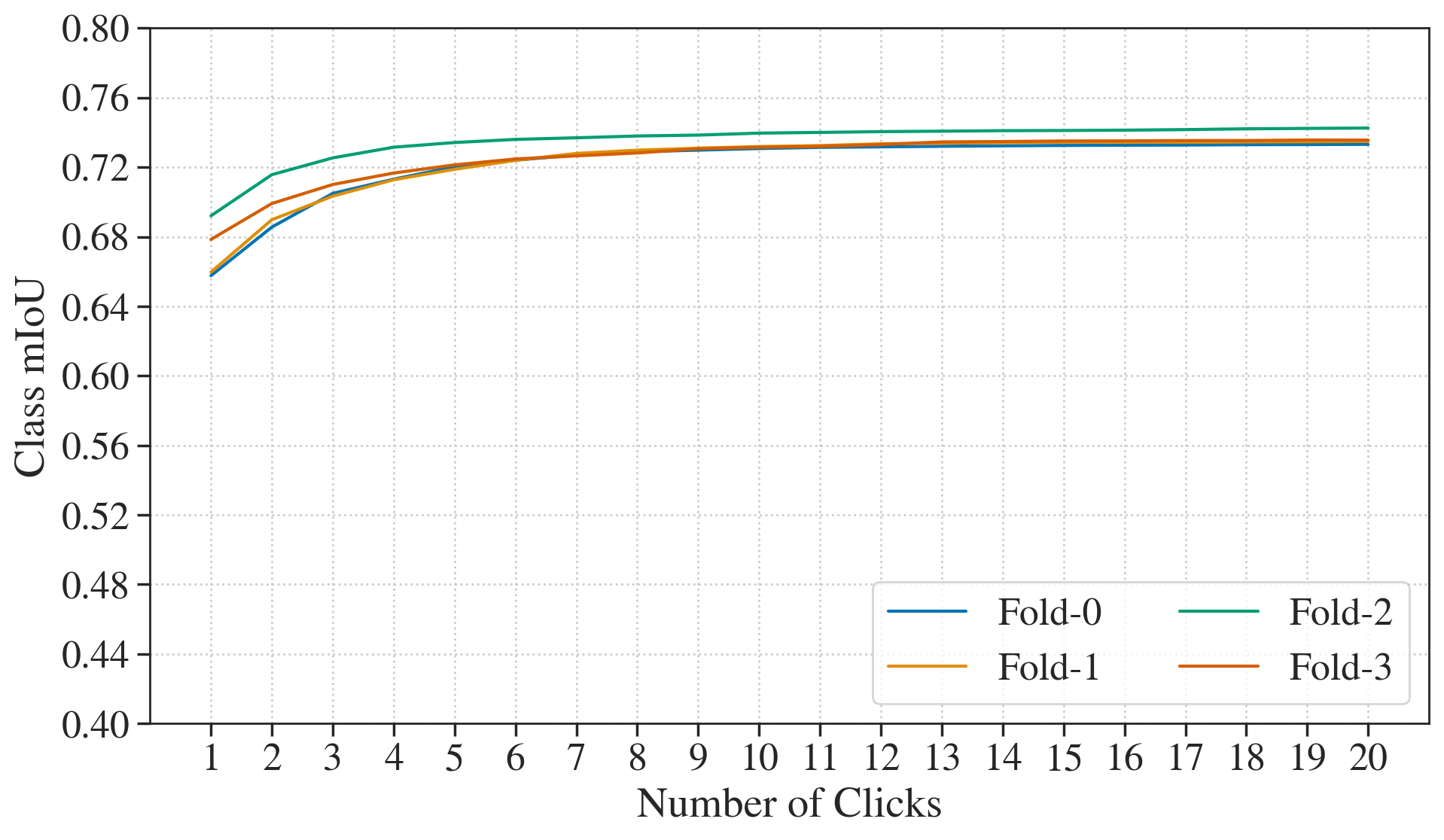}
    \caption{Results on query predictions for our 1-shot model trained on Pascal-$5^i$ for validation classes (upper panel) and training classes (lower panel).}
    \label{fig:query_results}
\end{figure}

We first present key results for query image predictions on the Pascal-$5^i$ dataset in Figure \ref{fig:query_results}. The plot shows Class mIoU, indicating improved performance with support image clicks. Notably, Split 1 exhibits higher IoUs on validation classes due to easier class composition.

We compare the performance of our 1-shot and 5-shot models with previous works in few-shot segmentation on the validation classes of different folds. Table \ref{table:query_results} demonstrates that our model achieves comparable results using sparse support information in the form of clicks, unlike previous works that rely on dense ground-truth support masks.

\begin{figure}[!ht]
    \centering
    \includegraphics[width=1.0\linewidth]{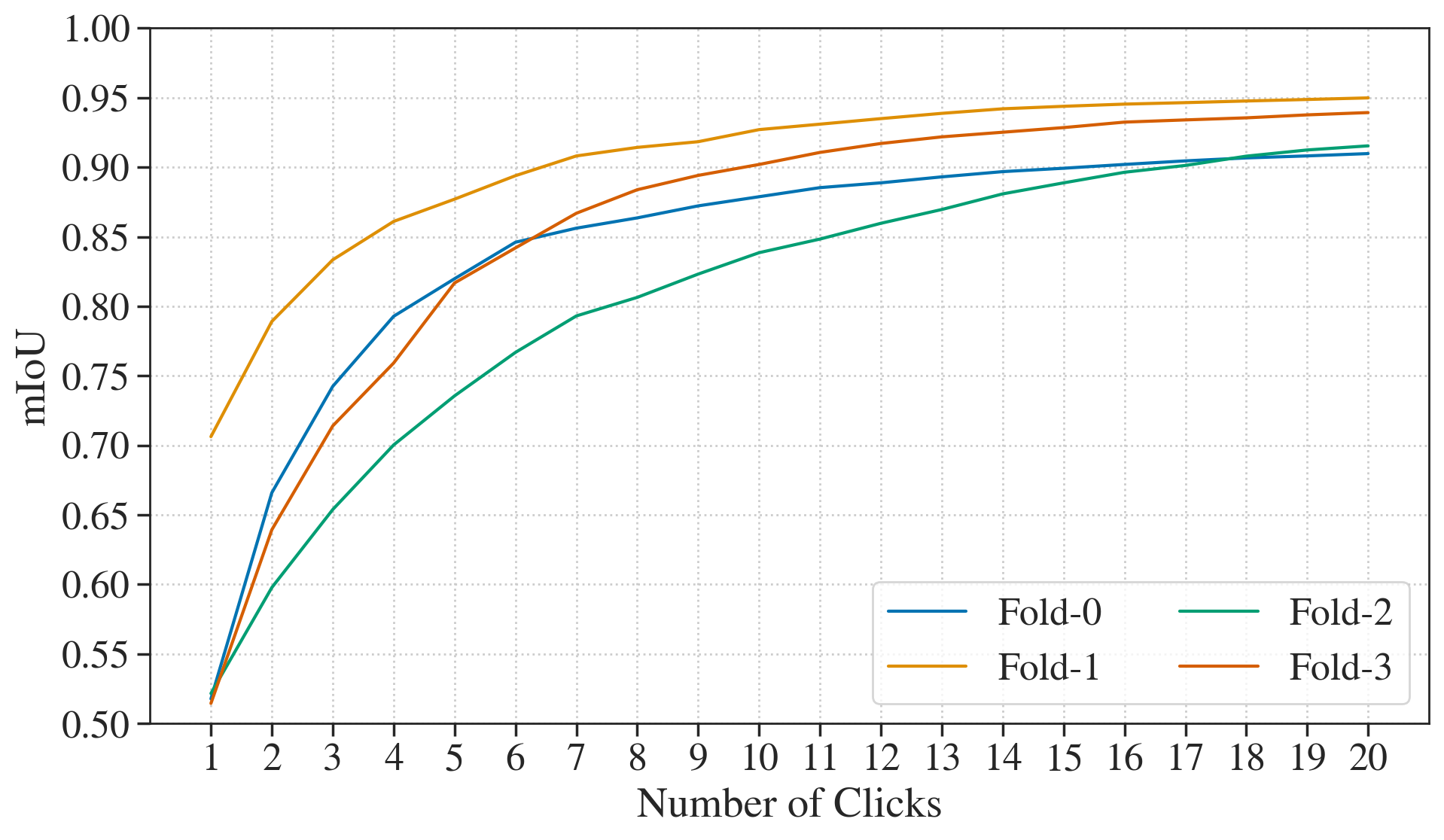}
    \includegraphics[width=1.0\linewidth]{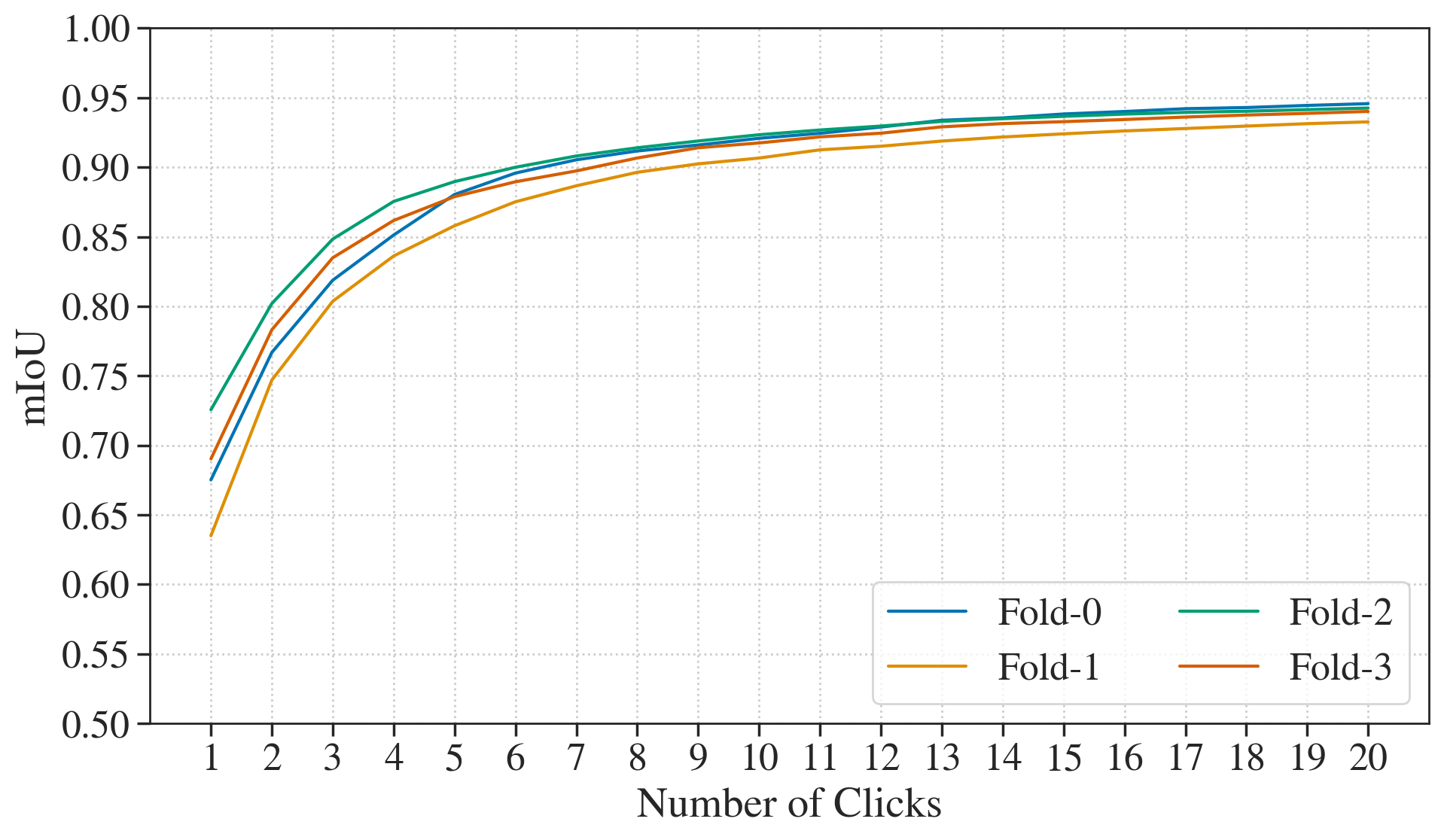}
    \caption{Results on support predictions for our 1-shot model trained on Pascal-$5^i$ for validation classes (upper panel) and training classes (lower panel).}
    \label{fig:support_results}
\end{figure}

In Figure \ref{fig:support_results}, we present mIoU results for support image predictions on the Pascal-$5^i$ dataset. Additional clicks have a more significant impact on the performance of support images compared to query images. The performance on validation classes eventually catches up with training classes, indicating that the model learns a meaningful understanding of clicks rather than relying solely on training class features.

\begin{table}
    \centering
    \begin{tabular}{cccc}
    \toprule[1.0pt]
    
    \multicolumn{2}{l}{Method} & NoC@85 & NoC@90 \\
    
    \toprule[1.0pt]
    
    \multicolumn{2}{l}{Latent diversity\cite{b5}} & 7.41 & 10.78 \\
    
    \multicolumn{2}{l}{BRS\cite{b4}} & 6.59 & 9.78 \\
    
    \multicolumn{2}{l}{f-BRS-B\cite{b12}} & 5.06 & 8.08 \\
    
    \multicolumn{2}{l}{RITM HRNet18 IT-M (SBD)\cite{b13}} & \textbf{3.39} & \textbf{5.43} \\
    
    \multicolumn{2}{l}{RITM HRNet18 IT-M (C+L)\cite{b13}} & 3.80 & \underline{6.06} \\
    
    \multicolumn{2}{l}{Ours} & \underline{3.79} & 7.08 \\
    
    \toprule[1.0pt]
    
    \end{tabular}
    
    \caption{Comparison of our model with previous works on interactive segmentation using training classes of SBD. For RITM, the training dataset is mentioned in parenthesis. Best results are bold, second-best are underlined.}

    \label{table:support_results}
    
\end{table}

We also compare our support module's mean performance on the training classes across different folds of Pascal-$5^i$ with various previous works in interactive segmentation in table \ref{table:support_results}. The metric NoC@x denotes the mean number of clicks to achieve an image mIoU of x.

\section{Conclusion and Future Work}

We have presented IFSENet, a combined network that successfully makes predictions on support images with the help of clicks, and propagates the relevant information to segment query images without clicks for novel classes. The model shows significant iterative improvement with additional clicks provided. The support performance is comparable to previous state-of-the-art interactive segmentation architectures, even though we use a much simpler and lighter U-net styled architecture. Our query performance, using only sparse supervision in the form of support clicks, is comparable to previous few-shot architectures employing dense support masks.

As part of our future work, we aim to develop a high-quality GUI-based application, which aids the user in interacting with our model via clicks and runs our algorithm in the backend smoothly and efficiently. We also plan to extend our model to other image domains such as medical, satellite, etc.


\begin{thebibliography}{00}
\bibitem{b1} Liang-Chieh Chen, George Papandreou, Iasonas Kokkinos, Kevin Murphy, and Alan L Yuille. Deeplab: Semantic image segmentation with deep convolutional nets, atrous convolution, and fully connected crfs. IEEE transactions on pattern analysis and machine intelligence, 40(4):834–848, 2017.
\bibitem{b2} Ming-Ming Cheng, Victor Adrian Prisacariu, Shuai Zheng, Philip HS Torr, and Carsten Rother. Densecut: Densely connected crfs for realtime grabcut. In Computer Graphics Forum, volume 34, pages 193–201. Wiley Online Library, 2015. 
\bibitem{b3} Nanqing Dong and Eric P. Xing. Few-shot semantic segmentation with prototype learning. In British Machine Vision Conference, 2018.
\bibitem{b4} Won-Dong Jang and Chang-Su Kim. Interactive image segmentation via backpropagating refinement scheme. In Proceedings of the IEEE/CVF Conference on Computer Vision and Pattern Recognition, pages 5297–5306, 2019.
\bibitem{b5} Zhuwen Li, Qifeng Chen, and Vladlen Koltun. Interactive image segmentation with latent diversity. In Proceedings of the IEEE Conference on Computer Vision and Pattern Recognition, pages 577–585, 2018.
\bibitem{b6} JunHao Liew, Yunchao Wei, Wei Xiong, Sim-Heng Ong, and Jiashi Feng. Regional interactive image segmentation networks. In 2017 IEEE international conference on computer vision (ICCV), pages 2746–2754. IEEE, 2017.
\bibitem{b7} Jun Hao Liew, Scott Cohen, Brian Price, Long Mai, SimHeng Ong, and Jiashi Feng. Multiseg: Semantically meaningful, scale-diverse segmentations from minimal user input. In Proceedings of the IEEE/CVF International Conference on Computer Vision, pages 662–670, 2019.
\bibitem{b8} Zheng Lin, Zhao Zhang, Lin-Zhuo Chen, Ming-Ming Cheng, and Shao-Ping Lu. Interactive image segmentation with first click attention. In Proceedings of the IEEE/CVF conference on computer vision and pattern recognition, pages 13339–13348, 2020.
\bibitem{b9}Jonathan Long, Evan Shelhamer, and Trevor Darrell. Fully convolutional networks for semantic segmentation. In Proceedings of the IEEE conference on computer vision and pattern recognition, pages 3431–3440, 2015.
\bibitem{b10}Olaf Ronneberger, Philipp Fischer, and Thomas Brox. U-net: Convolutional networks for biomedical image segmentation. In Medical Image Computing and Computer-Assisted Intervention–MICCAI 2015: 18th International Conference, Munich, Germany, October 5-9, 2015, Proceedings, Part III 18, pages 234–241. Springer, 2015.
\bibitem{b11}Carsten Rother, Vladimir Kolmogorov, and Andrew Blake. ” grabcut” interactive foreground extraction using iterated graph cuts. ACM transactions on graphics (TOG), 23(3):309–314, 2004. 
\bibitem{b12} Konstantin Sofiiuk, Ilia Petrov, Olga Barinova, and Anton Konushin. f-brs: Rethinking backpropagating refinement for interactive segmentation. In Proceedings of the IEEE/CVF Conference on Computer Vision and Pattern Recognition, pages 8623–8632, 2020.
\bibitem{b13} Konstantin Sofiiuk, Ilya A Petrov, and Anton Konushin. Reviving iterative training with mask guidance for interactive segmentation. In 2022 IEEE International Conference on Image Processing (ICIP), pages 3141–3145. IEEE, 2022.
\bibitem{b14} Zhuotao Tian, Hengshuang Zhao, Michelle Shu, Zhicheng Yang, Ruiyu Li, and Jiaya Jia. Prior guided feature enrichment network for few-shot segmentation. IEEE transactions on pattern analysis and machine intelligence, 44(2):1050–1065, 2020.
\bibitem{b15} Jingdong Wang, Ke Sun, Tianheng Cheng, Borui Jiang, Chaorui Deng, Yang Zhao, Dong Liu, Yadong Mu, Mingkui Tan, Xinggang Wang, et al. Deep high-resolution representation learning for visual recognition. IEEE transactions on pattern analysis and machine intelligence, 43(10):3349–3364, 2020.
\bibitem{b16} Kaixin Wang, Jun Hao Liew, Yingtian Zou, Daquan Zhou, and Jiashi Feng. Panet: Few-shot image semantic segmentation with prototype alignment. In proceedings of the IEEE/CVF international conference on computer vision, pages 9197–9206, 2019.
\bibitem{b17} Jiajun Wu, Yibiao Zhao, Jun-Yan Zhu, Siwei Luo, and Zhuowen Tu. Milcut: A sweeping line multiple instance learning paradigm for interactive image segmentation. In Proceedings of the IEEE Conference on Computer Vision and Pattern Recognition, pages 256–263, 2014.
\bibitem{b18} Ning Xu, Brian Price, Scott Cohen, Jimei Yang, and Thomas S Huang. Deep interactive object selection. In Proceedings of the IEEE conference on computer vision and pattern recognition, pages 373–381, 2016. 
\bibitem{b19} Chi Zhang, Guosheng Lin, Fayao Liu, Rui Yao, and Chunhua Shen. Canet: Class-agnostic segmentation networks with iterative refinement and attentive few-shot learning. In Proceedings of the IEEE/CVF Conference on Computer Vision and Pattern Recognition, pages 5217–5226, 2019.
\bibitem{b20} Xiaolin Zhang, Yunchao Wei, Yi Yang, and Thomas S Huang. Sg-one: Similarity guidance network for one-shot semantic segmentation. IEEE transactions on cybernetics, 50(9):3855–3865, 2020.
\bibitem{b21} Hengshuang Zhao, Jianping Shi, Xiaojuan Qi, Xiaogang Wang, and Jiaya Jia. Pyramid scene parsing network. In Proceedings of the IEEE conference on computer vision and pattern recognition, pages 2881–2890, 2017.
\end{thebibliography}
\end{document}